\documentclass{article}
 
\usepackage{graphicx}
\usepackage{amsfonts}
\usepackage{amsmath}
\usepackage{epstopdf}
\newcommand{\Bem}[1]{}
\newcommand{\deleted}[1]{}
\newcommand{\inserted}[1]{{  #1 }}

\usepackage{amsmath,amssymb,amsthm}
\usepackage{amsfonts}
\newtheorem{definition}{Definition}
\newtheorem{theorem}{Theorem}
\newtheorem{example}{Example}
\newtheorem{lemma}{Lemma}

\begin{document}
\title{On Seeking Consensus Between Document Similarity Measures}
\author{Mieczys{\l}aw K{\l}opotek
\\ Institute of Computer Science, Polish Academy of Sciences\\ ul. Jana Kazimierza 5, 01-248 Warszawa, 
Poland
 \\ {klopotek@ipipan.waw.pl}}
  
\maketitle

\begin{abstract}
This paper investigates the application of   consensus  clustering and 
 meta-clustering  
to the set of all possible partitions of a data set.
We show that when using a "complement" of Rand Index as a measure of cluster similarity, 
the total-separation partition, putting each element in a separate set, is chosen.
\end{abstract}

\section{Introduction}\label{sec:intro}

It is a well-known phenomenon that for the same data set 
various clustering algorithms may produce different partitions. 
This is true both for objects described by continuous variables (like results of measurements) 
and  for ones described by discrete features (like documents treated as points in term space).
Consensus clustering and meta-clustering are two known techniques 
helping to select the best one among the competing partitions. 
It is also well known that by changing the geometry of the data space 
we may even obtain all possible partitions of the dataset.

In this paper we investigate which   partition
would be selected if we apply consensus clustering or meta-clustering 
to the set of all possible partitions.    
In particular we formulate  (in Section 
\ref{sec:consensus-theorem}) and prove (in Section \ref{sec:consensus-proof})
that,  using the so-called Rand Index as a measure of partition similarity, 
we obtain via  consensus clustering
the partition putting each element in a separate set (which we will call subsequently \emph{total-separation partition}) as a "consensus" between these partitions. 
In Section \ref{sec:consensus-implications} we discuss briefly practical lessons from this theorem. 
In Section \ref{sec:consensus-limited-k} we demonstrate 
that a similar theorem can be formulated for the more realistic case where we consider only partitions containing not more clusters than a predefined threshold. 
In Section \ref{sec:metaclustering}
we show experimentally that also  
the very same similarity measure  
applied in   meta-clustering%
\footnote{Rand Index 
is frequently used in both consensus clustering and meta-clustering in search for a compromise clustering}
leads towards a similar choice of best partition.  

We start with Section \ref{sec:previous-work} explaining the concepts of concensus clustering and meta-clustering as well as pointing to the research on these topics. 
In Section \ref{sec:conclusions} we summarise our findings and point to further research directions.

\section{Previous work}\label{sec:previous-work}
Two main types of methods for handling the grieving issue of conflicting partitions of the same set  are currently under development 
in the literature:
\begin{itemize}
\item 	the meta-clustering
\item	the consensus clustering (called also in various brands ensemble clustering \cite{Wang:2011} or cluster aggregation \cite{Gionis:2007})
\end{itemize}

In the \emph{meta-clustering} stream   it is claimed that maybe the choice 
 of the best partition  should be left to the user
  who should only be assisted by grouping potential partitions into groups of similar ones.
To facilitate user selection of the right clustering,  
\cite{Caruana:2006}  (also  
compare  
\cite{Niu:2010,Bifulco:2009,Bifulco:2009b,Dasgupta:2010,Cui:2010}%
) 
  suggests to provide the user with meta-clusters (clusters of 
partitions) in order that the user better understands the choices. 
To facilitate their creation, \cite{Caruana:2006} proposes to use
a dissimilarity measure that they call  "Cluster Difference", closely related to 
 Rand Index (which is a similarity measure)  as  
  distance measure between partitions. 
 The sum of Rand Index and Cluster Difference is equal 1
The Rand Index (and as a consequence Cluster Difference) essentially is based on the calculation how many times elements are in the same or in different clusters.
Assume that the set $\mathbb{X}$ to be clustered, of cardinality $n$ consists of elements 
 $\{1,2, \dots, n\}$. 
Let   the quantity $ I_ {ij} $ be equal to 1 if the elements (objects)   $ i $ and $ j $ are in the same cluster of the first clustering, and different in the second one or vice versa. Otherwise, $ I_ {ij} $ is 0. Then the distance $CD$ between the two partitions $\Gamma_1,\Gamma_2$ (Cluster Difference) is defined as 
\begin{equation} \label{eq:CD} 
 CD(\Gamma_1,\Gamma_2)=\frac{\sum_{i, j\in \mathbb{X}, i <j} I_{ij}} {n (n-1) / 2} 
\end{equation}
where $ n $ is the number of objects in the collection. 
The value of $CD$ ranges from 0 for completely identical partitions 
to $1$ for extremely different ones. 
The extremes are e.g. two partitions:
all-in-one ($\Gamma$ consisting of exactly one cluster containing all elements)
and 
total-separation (every element in a separate cluster). 

If we wish to compare only partitions   over the same set of elements (with cardinality $n$), we can use the unnormalised version of $CD$:
\begin{equation} \label{eq:unCD} 
 unCD(\Gamma_1,\Gamma_2)= \sum_{i, j \in \mathbb{X}, i <j} I_{ij}  
\end{equation} 
In such a case the $unCD$ between all-in-one and total-separation partitions 
amounts to $\frac{n(n-1)}{2}$. 

But 
  here we encounter the problem: $n$ objects can be divided into $k$ clusters 
in $O(k^n)$ ways%
\footnote{ 
More precisely, \cite{CLU:And73} 
 shows that this number amounts to 
\begin{equation} 
\frac{1}{k!}\sum_{j=1}^{k}(-1)^{k-j}\Big(\begin{array}{l}k\\j\end{array}\Big)j^n \label{CLU:eq-liczebnosc} 
\end{equation} 
}.    
So we have to do with an NP-hard task.

In \emph{consensus clustering}  
\cite{Strehl:2003}
a kind of optimisation problem  (combinatorial optimisation) is formulated and solved. 
A similarity measure between partitions is introduced and data is re-clustered to get a clustering close to the original ones.
Alternatively groups of clusters are formed (a kind of meta-clustering) where 
the meta-clusters compete for objects performing just a re-clustering.
A number of other techniques in this direction was reviewed in 
\cite{Goder:2008:ALENEX,Hore:2009,Ghosh:2011:clensembles,Li:2008:weightedconsensus,Punera:2008,Monti:2003,Topchy:2005:clusteringensembles,Nguyen:2007,Wang:2014,ConsensusClusteringVogelNaumannDINA:2014}. 
We are particularly interested in the one 
initiated by \cite{Barthelemy:1995,Gordon:1998},
and further studied \cite{Goder:2008consensus} and applied  
\cite{Morlini:2010}.
This approach to  
consensus clustering seeks to find a new clustering that is as close as possible to all the partitions obtained.
The closeness is estimated e.g. (again) as the averaged Rand Index  \cite{Morlini:2010}.

\inserted{
It   has been noticed in the past that various clustering quality indices 
are biased, and in particular the Rand Index which we discuss in this paper,  see for example \cite{GroundTruthBiasLei:2016,Milligan:1986,Fowlkes:1983}. 
It has been reported that Rand Index tends to prefer smaller clusters rather than bigger ones when using it as an external cluster validity index, as for example in the study \cite{Fowlkes:1983}, which concentrated on clusterings into the same number of clusters. 
\cite{Fowlkes:1983} demonstrates both theoretically and empirically, that with the increase of the number of clusters, Rand index quickly heads towards stating that compared clusterings are identical.  
However, in this paper we do not handle the case of clusterings into one fixed number of clusters but rather allow for any number of clusters in a partition. 

\cite{GroundTruthBiasLei:2016} demonstrates, when studying balanced partitions with different number of clusters, that this time Rand index behaves differently, thus invalidating the generalisations from \cite{Fowlkes:1983}. The direction of the bias depends on the ground truth clustering (see Theorem 1 in \cite{GroundTruthBiasLei:2016} and further ones). The reversal of the bias was also reported from empirical studies \cite{Milligan:1986}. 
So in fact, the bias tendency of Rand Index remains under these circumstances undecided if we do not know the ground truth partition or we do not know whether it is balanced or not.  
\footnote{
Note that the bias has also been studied in the context 
of stable level of the indicator for randomly assigned partitions under various conditions, see e.g. 
\cite{Romano:2014}.
} 
Note that  the paper \cite{GroundTruthBiasLei:2016} and other  concentrate on the relationship between an empirical partition and the ground truth. 

However,  a study on Rand index bias in the context of consensus clustering and meta-clustering
seems not  have   been performed so far.  In such a case, the ground truth cannot be referred to  because it is not available.  \footnote{One may say  that we use Rand Index rather as internal and not external quality measure. }
Furthermore, the previous studies concentrate on "preferences" of Rand Index without questioning the existence of ground truth. We demonstrate here that it points at a clustering even if there is no ground truth, no really discernible base clustering.
}

\section{Theorem on consensus clustering}\label{sec:consensus-theorem}

It is well known that when we treat all possible geometries of the data set,
we can obtain all possible partitions of the data set.
Which one shall we choose? 
Under these circumstances, 
as we will  show,   consensus clustering is 
not 
useful in the selection process  of   the partition that is the closest to all the other
 ones. 
Because the closest partition 
 is a 
partition  that puts   each element (object, document) into a separate cluster.

In particular we will prove the following  
\begin{theorem} \label{th:consensus} 
For any $n$ objects, among all partitions 
the partition where each object falls into a separate cluster
(called subsequently \emph{total-separation partition})
has the lowest average distance to the other partitions in terms of Cluster Difference $CD$. 
\end{theorem}

As we consider a fixed $n$, let us concentrate on the unnormalised version $unCD$.

Note first one of the most serious implications of this theorem:
We consider a world of all possible partitions so one might think that this world is totally symmetric, and any partition may be a centre of such a universe. 
But this is not the case. 
The distance function distinguishes one of the elements.
So in fact it is biased in some way. 

There exist  plenty of other clustering quality assessment functions
and a similar analysis should be performed for them. 

So let us state beforehand that 
when performing a clustering task, 
we shall pick   at random neither the clustering function, nor the distance function nor the quality assessment function because each of them is biased and we shall care whether or not each function reflects our business purposes.

\section{The proof}\label{sec:consensus-proof}

 It is assumed that all the elements (objects, documents) of a set to be partitioned
possess identifiers being consecutive numbers starting with 1.
The proof will be performed by induction
on relabelling the objects in a cyclic manner 
combined with narrowing the set of candidates for the closest element.  
 
First step of checking validity of the theorem (subsection \ref{sec:consensus-proof-step0}) for small $n=2,3$ is trivial, but still necessary.
Subsequent subsections seek to establish the induction step by demonstrating a special role of the so-called simple extension of a partition  (to be introduced in subsection \ref{sec:consensus-proof-reducts}, along with the concept of reduct). 
The important feature here is that 
total-separation partition of $n+1$ elements 
is a simple extension 
of total-separation partition of $n$ elements. 
 Furthermore, as demonstrated in subsection \ref{sec:distred}, 
distances between partitions of $n+1$ elements
can be derived from distances between their reducts. 

The idea of the inductive step is as follows.
First we consider all extensions of a single partition. 
It is shown in subsection \ref{sec:extsinglepart} 
that on average the simple extension is the closest one to all the other extensions.

Then in subsection \ref{sec:disttoextofotherpart} we establish,
that among all extensions of one partition 
the simple extension is on average the closest one to all the extensions of some other partition. 

These two facts mean that among all extensions of a partition of a set of $n$ elements 
the simple extension is on average the closest one to all the partitions of a set of $n+1$ elemets. 
Hence, when looking for the consensus partition among all partitions of a set of $n+1$ elements  we need to consider only those partitions that are simple extensions of partitions of $n$ elements. 
In subsection \ref{sec:bestofsimpleext} we prove that 
among these candidates (the simple extensions) 
the simple extension of total-separation partition of $n$ elements,
that is total-separation partition of $n+1$ elements is the closest one on average. 
The induction proceed as follows. 
First we establish that the  solution (the partition closest to all) is among those partitions that have the $(n+1)$st element in a singleton cluster and at least 0 first elements being in  singleton clusters.
Then we have the inductive step:
If the solution is among 
those partitions that have the $(n+1)$st element in a singleton cluster and at least $i$ first elements being in  singleton clusters,
then 
  the solution is among 
those partitions that have the $(n+1)$st element in a singleton cluster and at least $i+1$ first elements being in  singleton clusters.
After $n$ inductive steps all $n+1$ elements will be in singleton clusters that is we get the total-separation partition as the solution.

\subsection{Cases $n=2,3$} \label{sec:consensus-proof-step0}
 
Consider the unnormalised version of Cluster Difference.

If $n=2$, then there exist only two partitions:
$\Gamma_{1;2}=\{\{1\},\{2\}\}$ and
$\Gamma_{2;2}=\{\{1,2\}\}$
The $unCD$ (as well as $CD$)   between them equals 1. 
So for both the average is identical and minimal.
The theorem is O.K.

With $n$=3 we get 
partitions 
\begin{itemize}
\item
$\Gamma_{1;3}=\{\{1\},\{2\},\{3\}\}$ (average $unCD$ distance  to other partitions  \textbf{1.5}, (normalised $CD$  0.5),  
\item
$\Gamma_{2;3}=\{\{1,3\},\{2\}\}$  (average $unCD$ distance  to other partitions 1.75)  
\item
$\Gamma_{3;3}=\{\{1\},\{2,3\}\}$  (average $unCD$ distance  to other partitions 1.75)  
\item
$\Gamma_{4;3}=\{\{1,2\},\{3\}\}$  (average $unCD$ distance  to other partitions 1.75)  
\item
$\Gamma_{5;3}=\{\{1,2,3\}\}$  (average $unCD$ distance  to other partitions 2.25).
\end{itemize}

Theorem is also in this case O.K. 
 
 \subsection{Case $n\rightarrow n+1$ - reducts and extensions} 
 \label{sec:consensus-proof-reducts} 
Consider now what happens when we have computed the unnormalised Cluster Difference  between partitions  for $n$ elements
and want to compute it for $n+1$ elements.

Each partition $\Gamma$ of $n+1$ elements has a unique partition $\Gamma^*$
 with $n$ elements (called its reduct) from which it can be derived
by adding the $(n+1)$st element to an existent cluster or by forming a new one.  $\Gamma$ on the other hand is called an \emph{extension} of $\Gamma^*$

Let us introduce the concepts of extension and reduct more formally.

\begin{definition}
Let $\Gamma^*$ be a partition of $n$ elements. 
\begin{itemize}
\item If $\Gamma$ is a partition of $n+1$ elements such that 
$\Gamma =\Gamma^*  \cup \{\{n+1\}\}$, then  $\Gamma$ will be called a \emph{ simple extension } of 
$\Gamma^*$ ($\Gamma =simpleextension(\Gamma^*)$)
\item If $\Gamma$ is a partition of $n+1$ elements such that 
there exists a set   $S\in \Gamma^*$ such that 
$\Gamma =\left(\Gamma^* - \{S\}\right) \cup \{S \cup \{n+1\} \}$, then    $\Gamma$ will be called a \emph{ complex extension } of 
$\Gamma^*$
\end{itemize}
Both simple extension and complex extension are \emph{extensions}. 
In both cases $\Gamma^*$ will be called a \emph{ reduct } of
$\Gamma$ ($\Gamma^* =reduct(\Gamma)$).
With $allextensions(\Gamma^*)$ we shall denote the set of all (simple and complex) extensions of $\Gamma^*$. 
\end{definition}
We distinguish complex and simple extensions in order to emphasise the role of a simple extension among the extensions of a partition -- on the one hand due to the simplicity of derivation of distances between extensions from the distances between their reducts (subsection \ref{sec:distred}), and on the other because a simple extension is closer to all the other extensions of the same partition than any complex extension, as will be shown later in Section \ref{sec:extsinglepart}. 

\begin{figure}
\includegraphics[width=0.8\textwidth]{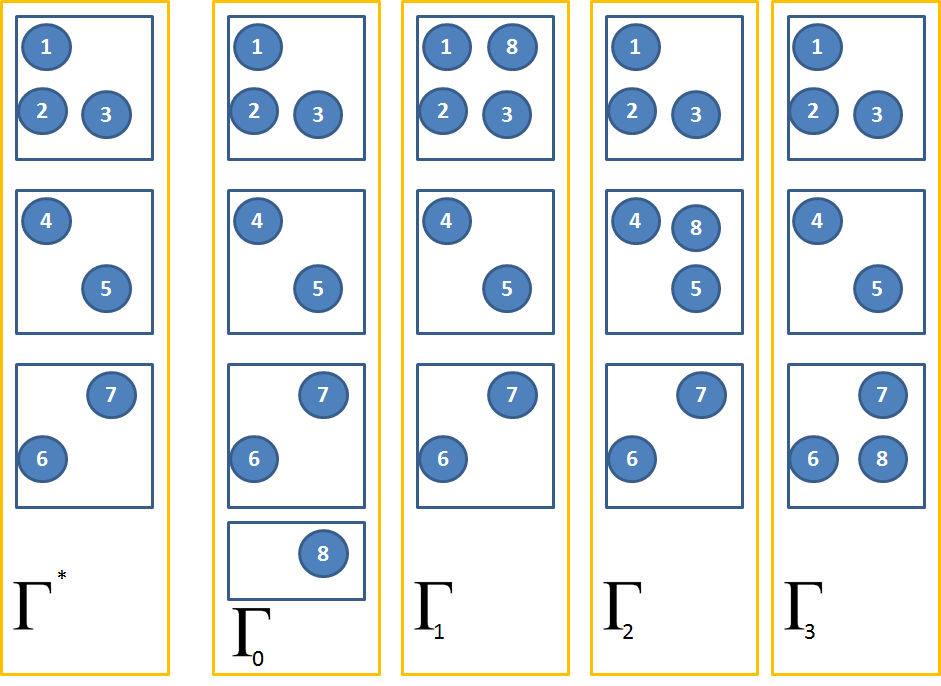}  %
\caption{Illustration of the concept of reducts,  simple and complex extensions.
The partition $\Gamma^*$ = 
$\{ \{1, 2, 3\} \{4, 5\} \{6, 7\}\}$
is a reduct for each of the partitions  
$\Gamma_0$=$  \{ \{1, 2, 3\} \{4, 5\} \{6, 7\} \{8\}\}$,
$\Gamma_1$=$ \{ \{1, 2, 3, 8\} \{4, 5\} \{6, 7\}\}$, 
$\Gamma_2$=$  \{ \{1, 2, 3\} \{4, 5, 8\} \{6, 7\}\}$ and
$\Gamma_3$=$  \{ \{1, 2, 3\} \{4, 5\} \{6, 7, 8\}\}$. 
$\Gamma_1, \Gamma_2, \Gamma_3$  are complex extensions of $\Gamma^*$.
$\Gamma_0$ is a simple extension of $\Gamma^*$.  
}\label{fig:reductextesions}
\end{figure}

\begin{example}{}
It is easily seen, using the notation from the previous section that
$\Gamma_{1;3},\Gamma_{2;3}, \Gamma_{3;3}$ are extensions of $\Gamma_{1;2}$,
while 
$\Gamma_{4;3},\Gamma_{5;3}$ are extensions of $\Gamma_{2;2}$.
$\Gamma_{1;3},\Gamma_{4;3}$ are hereby simple extensions, while
$\Gamma_{2;3}, \Gamma_{3;3}, \Gamma_{5;3}$ are complex extensions. 
\end{example}
For another example of a reduct, simple and complex extensions see Figure \ref{fig:reductextesions}.

 \subsection{Distances between partitions and their reducts} \label{sec:distred}

So consider  partitions $\Gamma_1,\Gamma_2$  of $n+1$ elements, being extensions of 
 two   partitions $\Gamma_1^*,\Gamma_2^*$ of $n$ elements resp. 
 Let us denote with $unCD(\Gamma_1^*,\Gamma_2^*;n)$ 
the unnormalised Cluster Difference  $unCD(\Gamma_1^*,\Gamma_2^*)$, 
where the parameter $n$ draws our attention to the fact that both partitions are defined over the set $\{1,...,n\}$.

Note that 
\begin{align} \label{eq:unCDnplus1} 
 unCD(\Gamma_1,\Gamma_2;n+1)
&= \sum_{i=1}^{n} \sum_{j=i+1}^{n+1}  I_{ij} 
 \\&= \sum_{i=1}^{n} \sum_{j=i+1}^{n}  I_{ij} +\sum_{i=1}^{n}I_{i,n+1}
\nonumber \\&= \sum_{i=1}^{n-1} \sum_{j=i+1}^{n}  I_{ij} +\sum_{i=1}^{n}I_{i,n+1}
\nonumber \\&=   unCD(\Gamma_1^*,\Gamma_2^*;n)+\sum_{i=1}^{n}I_{i,n+1}
\nonumber
\end{align}

This implies that 
if both $\Gamma_1,\Gamma_2$   are   simple extensions of 
$\Gamma_1^*,\Gamma_2^*$ that is \Bem{a \change{'}{*} was done}
$\Gamma_1=\Gamma_1^* \cup \{\{n+1\}\}$ 
and  $\Gamma_2=\Gamma_2^* \cup \{\{n+1\}\}$, then the unnormalised Cluster Difference between 
$\Gamma_1$ and $\Gamma_2$ is the same as between $\Gamma_1^*$ and $\Gamma_2^*$
because both in $\Gamma_1$ and $\Gamma_2$ element $n+1$ is separated from any other element. 

If $\Gamma_1=\Gamma_1^* \cup \{\{n+1\}\}$ and 
 $\Gamma_2^*-\Gamma_2=\{S_2\}$ where $S_2$ is not empty, then 
$ unCD(\Gamma_1,\Gamma_2;n+1)
=   unCD(\Gamma_1^*,\Gamma_2^*;n)+card(S_2)$.

If $\Gamma_1^*-\Gamma_1=\{S_1\}$ and 
 $\Gamma_2^*-\Gamma_2=\{S_2\}$ where $S_1,S_2$ are both  not empty, then 
the distance   
$ unCD(\Gamma_1,\Gamma_2;n+1)
=   unCD(\Gamma_1^*,\Gamma_2^*;n)+card(S_2-S_1)+card(S_1-S_2)$. 
\subsection{Centricity of a simple extension among all extensions}\label{sec:extsinglepart}

In this proof extensions of a partition play a very special role, because they constitute units for which properties of cumulative distances can be derived in closed form. In particular we show in this subsection that among the extensions of a partition the simple extension is the closest one to all the other extensions. In the next subsection we will demonstrate a similar property between the extensions of two different partitions. The formulas of this subsection are in fact special cases of those in the next subsection, but we believe that by separation of these cases the derivations will be easier to understand.

Let us now consider all extensions with $n+1$ elements 
of a partition $\Gamma^*$ of $n$ elements.
Let $\Gamma^*$  contain $k$ clusters $S_1,...,S_k$.
$\Gamma_0$ be the simple extension 
and $\Gamma_l$ be a complex extension 
containing the cluster $S_l\cup \{n+1\}$. 

Let us compute the sum of distances of simple extension to all the other extensions.
\begin{equation}
\sum_{l=1}^{k} unCD(\Gamma_0,\Gamma_l)
=\sum_{l=1}^{k} card(S_l)=n
\label{simpleall}
\end{equation}
Let us derive the formula for the sum of distances of a complex extension $\Gamma_{l'}$ to all the other extensions (remember that clusters of a partition are disjoint, $S_0$ be an empty set).
\begin{align}
\sum_{l=0,l\ne l'}^{k} unCD(\Gamma_l,\Gamma_{l'})
  &=\sum_{l=0,l\ne l'}^{k} \left(card\left(S_l-S_{l'} \right)+card\left(S_{l'}-S_l\right)\right)
\nonumber\\ &=\sum_{l=0,l\ne l'}^{k} \left(card\left(S_l \right)+card\left(S_{l'}\right)\right)
\nonumber\\ &= k \cdot card\left(S_{l'} \right) +\left(n-card\left(S_{l'} \right)\right) 
\nonumber\\ &=\left(k-1\right)\cdot card\left(S_{l'} \right) +n \ge n 
\label{eq:complexall}
\end{align}
%
%
Obviously, as the cardinality  of the set of extensions is fixed, 
the simple extension has the lowest average distance to other extensions among extensions of the same reduct. 

\begin{example}
Consider the partition $\Gamma^*=\{ \{1, 3\} \{2\}\}$ and its extensions.
The sums of distances of each of them to all the other are:
 5 for $ \{ \{1, 3, 4\} \{2\}\}$,
 4 for $ \{ \{1, 3\} \{2, 4\}\}$, 
 3 for $ \{ \{1, 3\} \{2\} \{4\}\}$.
The last one is the simple extension and has the lowest sum of distances. 
\end{example}

\subsection{Distance from a simple and a complex extension}\label{sec:disttoextofotherpart}
Let us now consider 
a simple extension $\Gamma_0$ and a complex one $\Gamma_m$, $m>0$ having the same reduct $\Gamma^*$
and all the partitions   $\Gamma'_l$ 
  with a different common reduct $\Gamma^*{}'$. 
Assume that both $\Gamma^*{}$ and $\Gamma^*{}'$ are partitions over the set $\{1,...,n\}$. 

Let $\Gamma^*{}'$  contain $k$ clusters $S'_1,...,S'_k$.
$\Gamma_0$ be the simple extension 
and $\Gamma_l$ be a complex extension 
containing the cluster $S_l\cup \{n+1\}$. $S_l$ defined as previously 

Let us calculate the sum of distances of simple extension $\Gamma_0$ to all the   extensions of $\Gamma^*{}'$.
\begin{align}
 \sum_{l=0}^{k} unCD(\Gamma_0,\Gamma'_l)
 &=\sum_{l=0}^{k} 
 \bigl( unCD\left(\Gamma^*,\Gamma^*{}';n\right)+card\left(S'_l\right)\bigr)
\nonumber\\&=n+\sum_{l=0}^{k} unCD(\Gamma^*,\Gamma^*{}';n) 
\nonumber\\&=n+(k+1)  unCD(\Gamma^*,\Gamma^*{}';n) \label{eq:simpleforeignall}
\end{align}

Let us determine the sum of distances of complex extension $\Gamma_m$ to all the   extensions of $\Gamma^*{}'$.
\begin{align}
\sum_{l=0}^{k} unCD(\Gamma_m,\Gamma'_l)
&=\sum_{l=0}^{k} 
 \left( unCD\left(\Gamma^*,\Gamma^*{}';n\right) +card\left(S_m-S'_l\right)+card\left(S'_l-S_m\right)\right)
\nonumber \\&=\sum_{l=0}^{k} 
 \left( unCD\left(\Gamma^*,\Gamma^*{}';n\right)\right.
+card\left(S_m\right)-card\left(S_m\cap S'_l\right)
\nonumber \\&\left.+card\left(S'_l\right)-card\left(S'_l\cap S_m\right)\right)
\nonumber \\&=\sum_{l=0}^{k} 
  unCD(\Gamma^*,\Gamma^*{}';n)
+(k+1)\cdot card(S_m)+n 
\nonumber \\&-\sum_{l=0}^{k}card(S_m\cap S'_l) -\sum_{l=0}^{k}card(S'_l\cap S_m)
\nonumber \\
&=\left(k+1\right) 
  unCD\left(\Gamma^*,\Gamma^*{}';n\right)+\left(k+1\right) \cdot card\left(S_m\right)+n 
\nonumber \\&-card\left(S_m\right) -card\left(S_m\right) 
\nonumber \\
&=(k+1) 
  unCD(\Gamma^*,\Gamma^*{}';n)+\left(k-1\right) \cdot card(S_m)+n 
\nonumber \\&\ge 
(k+1) 
  unCD(\Gamma^*,\Gamma^*{}';n)+n \label{eq:complexforeignall}
\end{align}

 
 Obviously, as the number of elements in the set of extensions  is fixed, 
the simple extension of $\Gamma^*$ has the lowest average distance to those extensions of  $\Gamma^*{}'$  
among extensions of the  $\Gamma^*$.

\begin{example}
Consider the partitions $\boldsymbol{\Gamma}=allextensions(\{ \{1, 3\} \{2\}\})$  
and let $\boldsymbol{\Gamma'}=allextensions(\{ \{1, 2\} \{3\}\})$.
Let us compute for each partition from the set $\boldsymbol{\Gamma}$ 
the sum of distances  to all partitions from $\boldsymbol{\Gamma'}$.
We obtain: 
 11  for $ \{ \{1, 3, 4\} \{2\}\}$,
 10  for $ \{ \{1, 3\} \{2, 4\}\}$,
 9  for $ \{ \{1, 3\} \{2\} \{4\}\}$.
The last partition is the simple extension and has the lowest sum of distances. 
\end{example}

Therefore we can conclude that if among the extensions 
of $\Gamma^*$ there exists a partition that is on average  the   closest  one  to any other partition, then this partition is for sure the simple extension of $\Gamma^*$.

 \subsection{
 The partition closest to all the others \Bem{s inserted}
}\label{sec:bestofsimpleext}

So we can summarize   sections \ref{sec:extsinglepart} and \ref{sec:disttoextofotherpart} as follows:
\begin{lemma}\label{lem:bestpart}
A partition of $n+1$ elements 
closest (on average) to all the other partitions
is  among the 
simple extensions of all the partitions of $n$ elements.
\end{lemma}
We can strengthen Lemma \ref{lem:bestpart} by stating:
\begin{lemma} \label{lem:bestpartfromset}
If among extensions of partitions of $n$ elements in the set $ \boldsymbol{\Gamma}$
there is a  partition of $n+1$ elements 
closest (on average) to all the other partitions of $n+1$ elements,
then such a partition 
exists among simple 
extensions of  $ \boldsymbol{\Gamma}$.
\end{lemma} 

Now we are ready to prove by induction that the total-separation partition is the closest on average to all partitions. 

Our working hypothesis is as follow:
For any $i=0,...,n$,
the solution of the problem of finding the partition of $n+1$ elements closest on average to all the possible partitions of the same set
is contained in the set $\boldsymbol{C_i}$ of such partitions for which 
the $(n+1)$st \Bem{changed} element constitutes a singleton in this partition (a cluster with one element only)
and the elements $1,2,...,i$ constitute also singletons. 
Obviously, if $i=n-1$, then the set containing the solution contains one element only that is the total-separation partition (if all elements but $n$ are singletons, then also  $n$ is). 

Let us first establish the validity of the claim for $i=0$. 
Let 
$ \boldsymbol{\Gamma_{A(n)}}$
denote the set of all  partitions of $n$ elements. 
Lemma \ref{lem:bestpart}  states that a partition closest on average to all partitions in $ \boldsymbol{\Gamma_{A(n+1)}}$ is among simple extensions of $ \boldsymbol{\Gamma_{A(n)}}$.
So this is exactly   the set of candidates for the on average closest elements denoted previously as   
$\boldsymbol{C_0}$. 
So initially 
$\boldsymbol{C_0}=simpleextension(\boldsymbol{\Gamma_{A(n)}})$, where $simpleextension()$ is a function producing simple extensions of partitions. 

Next we shall prove the inductive step. That is that if 
$\boldsymbol{C_i}$ contains the solution then $\boldsymbol{C_{i+1}}$ contains also the solution. 
For this purpose 
  consider the operation of re-labelling elements of partitions. 
If $\Gamma$ is a partition, then $relabel(\Gamma)$ is a partition obtained by changing  an identifier $i$ of an element to $i+1$ except for the element with the highest identifier $n+1$ that will be turned to $1$. 
It is obvious that $unCD\left(\Gamma_1,\Gamma_2\right)=unCD\left(relabel\left(\Gamma_1\right),relabel\left(\Gamma_2\right)\right)$. It is also obvious that $relabel(\boldsymbol{\Gamma_{A(n+1)}})=\boldsymbol{\Gamma_{A(n+1)}}$, though in general $relabel(\boldsymbol{C_i})\ne \boldsymbol{C_i}$. 
However, an element closest on average to each element of $\boldsymbol{\Gamma_{A(n+1)}}$ is among $relabel\left(\boldsymbol{C_i}\right)$. 
But note that 
$ simpleextension\left(reduct\left(relabel\left(\boldsymbol{C_i}\right)\right)\right) 
\subseteq relabel\left(\boldsymbol{C_i}\right)
$ (as we consider $n\ge3)$. 
Therefore according to Lemma \ref{lem:bestpartfromset} 
  we can obtain a new candidate set by the operation
$\boldsymbol{C_{i+1}}:=
simpleextension\left(reduct
			\left(relabel
				\left(\boldsymbol{C_i}
				\right)
			\right)
		\right)$.
This is because  in  $relabel\left(\boldsymbol{C_i}\right)$ in each partition elements $1,...,i+1$ are singletons, as $1,...,i$ were singletons in $\boldsymbol{C_i}$ as well as $n+1$ and now via relabeling they became $2,...,i+1$ and $1$ respectively.  
The operation $simpleextension\left(reduct\left(\right)\right)$ eliminates 
everything from $relabel\left(\boldsymbol{C_i}\right)$ except for simple extensions which have $n+1$ as singletons. 
This proves the validity of the induction step. 
Obviously, the set of candidates will be reduced in this way.

By induction our claim is valid. Theorem \ref{th:consensus} is proven.

\begin{example}
Consider the set 
\begin{align}
\boldsymbol{\Gamma_{A(4)}}=
 & \{\{ \{1, 2, 3, 4\}\} \ ,\Bem{\nonumber\\ 
 &}  \{ \{1, 2, 3\} \{4\}\}	\ ,\Bem{\ ,\nonumber\\  
 &}  \{ \{1, 2, 4\} \{3\}\}	\ ,\Bem{\nonumber\\ 
 &}  \{ \{1, 2\} \{3, 4\}\}	\ ,\nonumber\\ 
 &  \{ \{1, 2\} \{3\} \{4\}\}	\ ,\Bem{\nonumber\\ 
 &}  \{ \{1, 3, 4\} \{2\}\}	\ ,\Bem{\nonumber\\ 
 &}  \{ \{1, 3\} \{2, 4\}\}	\ ,\Bem{\nonumber\\ 
 &}  \{ \{1, 3\} \{2\} \{4\}\}	\ ,\nonumber\\ 
 &  \{ \{1, 4\} \{2, 3\}\}	\ ,\Bem{\nonumber\\ 
 &}  \{ \{1\} \{2, 3, 4\}\}	\ ,\Bem{\nonumber\\ 
 &}  \{ \{1\} \{2, 3\} \{4\}\}	\ ,\Bem{\nonumber\\ 
 &}  \{ \{1, 4\} \{2\} \{3\}\}	\ ,\nonumber\\ 
 &  \{ \{1\} \{2, 4\} \{3\}\}	\ ,\Bem{\nonumber\\ 
 &}  \{ \{1\} \{2\} \{3, 4\}\}	\ ,\Bem{\nonumber\\  
 &}  \{ \{1\} \{2\} \{3\} \{4\}\}	 \}  \nonumber
\end{align}
The set of candidates $\boldsymbol{C_0}$, being the simple extensions among the above, is 
\begin{align}
\boldsymbol{C_0}= &
 \{
 \{ \{1, 2, 3\} \{4\}\}	\ , \ 
 \{ \{1, 2\} \{3\} \{4\}\}	\ ,\ 
 \{ \{1, 3\} \{2\} \{4\}\}	\ ,\nonumber\\ &
 \{ \{1\} \{2, 3\} \{4\}\}	\ ,\ 
 \{ \{1\} \{2\} \{3\} \{4\}\}
\}\nonumber
\end{align} 
$relabel(\boldsymbol{C_0})$ changes it to
\begin{align}
relabel(\boldsymbol{C_0})= &
\{
 \{ \{1\} \{2, 3, 4\}\}	\ ,\ 
 \{ \{1\} \{2, 3\} \{4\}\}	\ ,\  
 \{ \{1\} \{2, 4\} \{3\}\}	\ ,\nonumber\\ & 
 \{ \{1\} \{2\} \{3, 4\}\}	\ ,\  
 \{ \{1\} \{2\} \{3\} \{4\}\}
\} \nonumber
\end{align} 
The transformation to 
$\boldsymbol{C_1}$ yields 
$\boldsymbol{C_1}=\{
 \{ \{1\} \{2, 3\} \{4\}\}	\ , \ 
 \{ \{1\} \{2\} \{3\} \{4\}\}
\}$. 
The transformation to 
$\boldsymbol{C_2}$ yields 
$\boldsymbol{C_2}=\{
 \{ \{1\} \{2\} \{3\} \{4\}\}
\}$. \Bem{periods inserted}
\end{example}

 \section{Practical implications}\label{sec:consensus-implications}

At the first glance Theorem \ref{th:consensus} may seem to be 
trivial, useless, unrealistic and  impractical. 
Trivial because it may appear to be obvious that a total-separation partition is closest to all the other partitions. 
Useless because nobody is interested in obtaining a consensus in terms of a  
total-separation partition.
Unrealistic because the space of all possible clusterings is so immense, that for real world sample sizes one would never run so many clustering algorithms  as to fill the whole partition universe. 
Impractical because one usually restricts the number of clusters $k$ in a partition by an upper bound (much) lower than the number of elements $n$. 

In order to demonstrate that these intuitions are wrong, we performed a series of simulation experiments results of which are summarized in Tables 
\ref{tab:randomconsensus} (sampling the universe, \inserted{no structure in the underlying data assumed}), \ref{tab:randomconsensusmodified} (sampling the universe with a modified partition dissimilarity measure, \inserted{no structure in the underlying data assumed}),  
\ref{tab:structconsensus} (sampling a subuniverse where the presence of a simple structure in the data is assumed), and further ones, discussed in the next section. 

\inserted{
With the experiments, we address  the following questions:
\begin{itemize}
\item Does the bias of Rand index to choose the total-separation partition 
as consensus for the whole universe of partitions persist if we consider only a uniform random sample from this universe?
\item Is this bias a general property of cluster quality indices 
or is it specific to Rand index?
\item Does the bias of Rand index to choose the total-separation partition 
pertain if there is a structure in the partitions for which a consensus is sought?  
\end{itemize}
While these questions are addressed in the current section, we pose them again in the next section in the context of constraining the set of partitions to those with an upper bound on the number of partitions.

Recall that 
  consensus clustering  uses 
  Cluster Difference (derived from Rand Index) as a measure of distance 
between partitions in order to identify the consensus partition. 
We have already demonstrated theoretically, that if the set of our clustering algorithms would yield all possible partitions, then Rand index would pick up the total-separation partition. 
But of course the space of all partitions is too large so that we will never get all possible partitions. 
Nonetheless by manipulating clustering algorithm parameters in an irresponsible way we can obtain a random sample from this universe.
In fact it is quite easy to invent clustering algorithms delivering for the same set of data any clustering we want.
This section simulates such a situation and shows experimentally what the outcome of consensus clustering will be questionable also in such a case. 
See comments on experiments in tables  \ref{tab:randomconsensus} and \ref{tab:randomconsensusmodified} below. 
This suggests that the user exploiting the technology of consensus clustering   must at least have an approximate  vision of the geometry of the data space and parametrise clustering algorithms in a way not disturbing this geometry.  
Only in this case the consensus clustering  may be helpful in the choice of appropriate compromise clustering. 
See the comments below on comparison of tables 
\ref{tab:randomconsensus} and 
\ref{tab:structconsensus}.

}

The experiments consisted in drawing 1,000 samples from the partition universe for each parameter setting (characterised by columns 1-4) and computed results are presented in column 5 (eventually column 6). 
So for example in Table \ref{tab:randomconsensus} in the second data row we have the information, that samples were drawn from the universe of partitions over 4 elements, that the number of possible partitions is 15, out of them samples of 8 partitions were drawn which were intended to be 50\% of the sample space, and the evaluation result was 88.2\%. 

The experiment underlying Table \ref{tab:randomconsensus} was devised to find out how often the total-separation partition will turn out to be the consensus partition for a uniformly randomly drawn sample of partitions. The results are visible in column 5.

The experiment underlying Table \ref{tab:randomconsensusmodified}  was essentially the same as the previous one, but instead of dissimilarity measure $unCD$ its modification $unCD_m(a,b)=unCD^{10}(a,b)\inserted{=\left(unCD(a,b)\right)^{10}}$ was used. Again the total-separation partition occurrence as consensus partition was counted.   The results are visible in column 5. 

The experiment underlying Table \ref{tab:structconsensus} 
differs from the previous two in that not the whole universe of partitions of $n$ elements was considered, but only those partitions for which there exists a "structure" that is where elements $1,3$ occur always in the same cluster. 
Here $unCD$ was used as distance measure as in Table \ref{tab:randomconsensus}.
The 5th column counts the number of occurrences of total-separation cluster as consensus, while the 6th column tells how frequently the expected partition (where all elements are singletons except for elements 1 and 3 that constitute one cluster) is discovered as consensus partition. 
 
Comparing Tables \ref{tab:randomconsensus} and \ref{tab:randomconsensusmodified} (rows where the 3rd column contains 100\% entry) we see first of all that the result of the Theorem 
\ref{th:consensus} is \emph{far from being trivial}. Total-separation partition does not need to be the default choice for a consensus of the entire universe of possible partitions. A skilled choice of dissimilarity function may point at any partition. It is the particular property of Rand Index (Cluster Difference) that distinguishes total-separation partition. So one can say that Rand Index is biased towards total-separation partition in case of no pattern in the data.  
\inserted{This property seems to be vital and has never been reported in the context of consensus clustering. An investigation of other measures with respect to their behaviour under missing structure should be carried out. 
}

\deleted{
When we compare tables \ref{tab:randomconsensus} and 
\ref{tab:structconsensus}, we see immediately that the property of being able to provide total-separation partition as the consensus is a very \emph{useful} property because it allows us to discern between a meaningful and meaningless set of outcomes of diverse clustering algorithms.
}
\inserted{
Let us compare column 5 of tables \ref{tab:randomconsensus} and 
\ref{tab:structconsensus}. 
In table  \ref{tab:randomconsensus} we simulated the case that there was no intrinsic structure behind the data, so that the partitions obtained from various algorithms just provide random samples from the universe of all possible partitions. 
We count how frequently the total-separation partition will occur as the consensus between the diverse partitions. 
It turns out that it happens quite often even if we have small sets of partitions. 
In table \ref{tab:structconsensus} on the other hand such a situation never happens. Furthermore, in column 6 of this table one can see that the centre of the set of clusters exhibiting the assumed structure occurs quite often as the consensus in this experiment. 
So non-occurring of total-separation partition as a consensus under a sufficiently large set of competing partitions  can be considered as a good indicator of existence of some structure in the data. 
So  the ability  of a consensus clustering algorithm   to provide total-separation partition as the consensus is a very \emph{useful} property because it allows us to discern between a meaningful and meaningless set of outcomes of diverse clustering algorithms.
}

The Theorem \ref{th:consensus} provides us with an important insight into the partition space because the indicated behaviours are observed not only in the whole universe, but also in sufficiently big samples. 
Hence the distance of the actual consensus from the total-separation partition is an important indicator of the actual structure in the data. 
Enforcing exclusion of total-separation partition from consensus seeking algorithm is not a wise decision. 
So the result is very \emph{practical}. 

Let us also stress that for too small sample sizes one can get impression of existence of a structure in the data even if there is none. Hence in practice one should verify the validity of the consensus in the application domain. 

We will postpone the discussion of the restriction  of the number of clusters $k$ in a partition by an upper bound to the next section as it requires some additional theoretical discussion. 
 
\begin{table}
\caption{Results of consensus clustering of randomly selected partitions}\label{tab:randomconsensus}
\begin{tabular}{|p{2cm}|p{2cm}|p{2cm}|p{2cm}|p{2cm}|}
\hline
data set size & number of all possible clusterings 
& sample size & percentage of possible clusterings & percentage of total-separation partitions in consensus \\
\hline
\textit{1} & \textit{2} & \textit{3} & \textit{4} & \textit{5} \\
\hline
\hline
4 & 15 & 15 & 100 \% & 100 \% \\ 
4 & 15 & 8 & 50 \% & 88.2 \% \\ 
\hline
5 & 52 & 52 & 100 \% & 100 \% \\ 
5 & 52 & 26 & 50 \% & 100 \% \\ 
5 & 52 & 21 & 40 \% & 97.2 \% \\ 
5 & 52 & 10 & 20 \% & 78.9 \% \\ 
5 & 52 & 5 & 10 \% & 23.1 \% \\ 
\hline
6 & 203 & 203 & 100 \% & 100 \% \\ 
6 & 203 & 102 & 50 \% & 100 \% \\ 
6 & 203 & 81 & 40 \% & 100 \% \\ 
6 & 203 & 41 & 20 \% & 99.9 \% \\ 
6 & 203 & 20 & 10 \% & 96.5 \% \\ 
\hline
7 & 877 & 88 & 10 \% & 100 \% \\ 
7 & 877 & 44 & 5 \% & 99.9 \% \\ 
7 & 877 & 35 & 4 \% & 99.7 \% \\ 
7 & 877 & 18 & 2 \% & 93.7 \% \\ 
7 & 877 & 9 & 1 \% & 49.9 \% \\ 
\hline
8 & 4140 & 41 & 1 \% & 100 \% \\ 
8 & 4140 & 21 & 0.5 \% & 94 \% \\ 
8 & 4140 & 17 & 0.4 \% & 89 \% \\ 
8 & 4140 & 8 & 0.2 \% & 61 \% \\ 
8 & 4140 & 4 & 0.1 \% & 38 \% \\ 
\hline
\end{tabular}
\end{table}

\begin{table}
\caption{Results of consensus clustering of randomly selected partitions for a modified distance measure}\label{tab:randomconsensusmodified}
\begin{tabular}{|p{2cm}|p{2cm}|p{2cm}|p{2cm}|p{2cm}|}
\hline
data set size & number of all possible clusterings 
& sample size & percentage of possible clusterings & percentage of total-separation partitions in consensus \\
\hline
\textit{1} & \textit{2} & \textit{3} & \textit{4} & \textit{5} \\
\hline
\hline
4 & 15 & 15 & 100 \% & 0 \% \\ 
4 & 15 & 8 & 50 \% & 30.2 \% \\ 
\hline
5 & 52 & 52 & 100 \% & 0 \% \\ 
5 & 52 & 26 & 50 \% & 6.5 \% \\ 
5 & 52 & 21 & 40 \% & 4.4 \% \\ 
5 & 52 & 10 & 20 \% & 16.2 \% \\ 
5 & 52 & 5 & 10 \% & 16.4 \% \\ 
\hline
6 & 203 & 203 & 100 \% & 0 \% \\ 
6 & 203 & 102 & 50 \% & 0.7 \% \\ 
6 & 203 & 81 & 40 \% & 3.2 \% \\ 
6 & 203 & 41 & 20 \% & 11.3 \% \\ 
6 & 203 & 20 & 10 \% & 5.5 \% \\ 
\hline
7 & 877 & 88 & 10 \% & 1.5 \% \\ 
7 & 877 & 44 & 5 \% & 2.3 \% \\ 
7 & 877 & 35 & 4 \% & 2.9 \% \\ 
7 & 877 & 18 & 2 \% & 8.6 \% \\ 
7 & 877 & 9 & 1 \% & 6.8 \% \\

\hline
\end{tabular}
\end{table}

\begin{table}
\caption{Results of consensus clustering of randomly selected partitions from a set of partitions exhibiting simple structure}\label{tab:structconsensus}
\begin{tabular}{|p{2cm}|p{2cm}|p{2cm}|p{2cm}|p{2cm}|p{2cm}|}
\hline
data set size & number of all  clusterings with structure
& sample size & percentage of possible clusterings & percentage of total-separation partitions in consensus & percentage of structure set centre \\
\hline
\textit{1} & \textit{2} & \textit{3} & \textit{4} & \textit{5} & \textit{6}\\
\hline
\hline

4 & 5 & 5 & 100 \% & 0 \% & 100 \% \\ 
4 & 5 & 2.5 & 50 \% & 0 \% & 69.2 \% \\ 
\hline
5 & 15 & 15 & 100 \% & 0 \% & 100 \% \\ 
5 & 15 & 8 & 50 \% & 0 \% & 89.7 \% \\ 
5 & 15 & 6 & 40 \% & 0 \% & 74 \% \\ 
5 & 15 & 3 & 20 \% & 0 \% & 22.8 \% \\ 
5 & 15 & 2 & 10 \% & 0 \% & 53.1 \% \\ 
\hline
6 & 52 & 52 & 100 \% & 0 \% & 100 \% \\ 
6 & 52 & 26 & 50 \% & 0 \% & 100 \% \\ 
6 & 52 & 21 & 40 \% & 0 \% & 97.2 \% \\ 
6 & 52 & 10 & 20 \% & 0 \% & 77.3 \% \\ 
6 & 52 & 5 & 10 \% & 0 \% & 23.5 \% \\ 
\hline

7 & 203 & 41 & 20 \% & 0 \% & 99.9 \% \\ 
7 & 203 & 20 & 10 \% & 0 \% & 95.8 \% \\ 
7 & 203 & 16 & 8 \% & 0 \% & 90.4 \% \\ 
7 & 203 & 8 & 4 \% & 0 \% & 70.2 \% \\ 
7 & 203 & 4 & 2 \% & 0 \% & 47.6 \% \\ 
 
\hline
\end{tabular}
\end{table}

\inserted{
In summary, the experiments allow to answer the posed questions as follows: 
\begin{itemize}
\item The bias of Rand index to choose the total-separation partition 
as consensus for the whole universe of partitions \emph{persists} if we consider only a uniform random sample from this universe?
\item This bias \emph{is not} a general property of cluster quality indices 
and the direction of the bias should be investigated separately for other cluster quality indices. 
\item  Rand index behaves differently  if there is a structure in the partitions for which a consensus is sought. 
\end{itemize}
}

\section{Imposing a limit on the number of clusters to be considered}\label{sec:consensus-limited-k}

Partitions explored  practically in the case of consensus clustering always contain 
much fewer  clusters compared to the size of the set of objects
$(k<<n)$. 
The question is: would a limitation on the number of clusters $k<k_{max}$
change anything with respect to the previous results?

Note that for a partition consisting of $k$ sets each of its complex extensions contains also $k$ sets, but the simple extension contains $k+1$ sets. Therefore, as long as $k<k_{max}$, nothing changes in the discussion of sections 
  \ref{sec:extsinglepart} and \ref{sec:disttoextofotherpart}.  But for $k=k_{max}$, we need to update 
equations
(\ref{eq:complexall}),
(\ref{eq:simpleforeignall}) and 
(\ref{eq:complexforeignall}), because we will no longer consider (hence count) distances to the simple extensions.
So the equation (\ref{eq:complexall}) has to be substituted by 

\begin{align}
\sum_{l=1,l\ne l'}^{k} unCD(\Gamma_l,\Gamma_{l'})
  &=\sum_{l=1,l\ne l'}^{k} \left(card\left(S_l-S_{l'} \right)+card\left(S_{l'}-S_l\right)\right)
\nonumber\\ &=\sum_{l=1,l\ne l'}^{k} \left(card\left(S_l \right)+card\left(S_{l'}\right)\right)
\nonumber\\ &= (k-1) \cdot card\left(S_{l'} \right) +\left(n-card\left(S_{l'} \right)\right) 
\nonumber\\ &=\left(k-2\right)\cdot card\left(S_{l'} \right) +n \ge n 
\label{eq:complexcomplex}
\end{align}
\noindent which holds of course only if $k_{max}>1$, which is rather a non-restrictive assumption. 

The equation (\ref{eq:simpleforeignall}) has to be substituted by 
\begin{align}
 \sum_{l=1}^{k} unCD(\Gamma_0,\Gamma'_l)
 &=\sum_{l=1}^{k} 
 \bigl( unCD\left(\Gamma^*,\Gamma^*{}';n\right)+card\left(S'_l\right)\bigr)
\nonumber\\&=n+\sum_{l=1}^{k} unCD(\Gamma^*,\Gamma^*{}';n) 
\nonumber\\&=n+k\cdot  unCD(\Gamma^*,\Gamma^*{}';n) \label{eq:simpleforeigncomplex}
\end{align}

The equation (\ref{eq:complexforeignall}) has to be substituted by 

\begin{align}
\sum_{l=1}^{k} unCD(\Gamma_m,\Gamma'_l)
&=\sum_{l=1}^{k} 
 \left( unCD\left(\Gamma^*,\Gamma^*{}';n\right) +card\left(S_m-S'_l\right)+card\left(S'_l-S_m\right)\right)
\nonumber \\&=\sum_{l=1}^{k} 
 \left( unCD\left(\Gamma^*,\Gamma^*{}';n\right)\right.
+card\left(S_m\right)-card\left(S_m\cap S'_l\right)
\nonumber \\&\left.+card\left(S'_l\right)-card\left(S'_l\cap S_m\right)\right)
\nonumber \\&=\sum_{l=1}^{k} 
  unCD(\Gamma^*,\Gamma^*{}';n)
+k\cdot card(S_m)+n 
\nonumber \\&-\sum_{l=1}^{k}card(S_m\cap S'_l) -\sum_{l=1}^{k}card(S'_l\cap S_m)
\nonumber \\&=k\cdot 
  unCD\left(\Gamma^*,\Gamma^*{}';n\right)+k\cdot card\left(S_m\right)+n 
\nonumber \\&-card\left(S_m\right) -card\left(S_m\right) 
\nonumber \\&=k\cdot
  unCD(\Gamma^*,\Gamma^*{}';n)+(k-2)\cdot card(S_m)+n 
\nonumber \\&\ge 
k\cdot
  unCD(\Gamma^*,\Gamma^*{}';n)+n \label{eq:complexforeigncomplex}
\end{align}
 which is again valid under $k_{max}>1$. 

Hence lemmas \ref{lem:bestpart} and \ref{lem:bestpartfromset} retain their validity for the restricting $k_{max}$ and hence the reasoning presented in section \ref{sec:bestofsimpleext}.

So the following theorem holds:

\begin{theorem} \label{th:consensusrestricted} 
For any $n$ objects, among all partitions 
the partition where each object falls into a separate cluster
(called subsequently \emph{total-separation partition})
has the lowest average Cluster Difference $CD$ to the other partitions
consisting of  at most $k_{max}>1$ clusters. 
\end{theorem}

But of course the limitation of the number of clusters still leaves a huge space of possible partitions of the set of elements. 
So let us return to the discussion of sampling this space.

Experiments analogous to those of previous section have been performed and are summarized in  Table \ref{tab:randomconsensusklimited}. Random samples from the space of permissable clusterings (with $k<k_{max}$\inserted{)} were drawn and the suitability of total-separation partition as consensus cluster was checked. A similar pattern to table \ref{tab:randomconsensus} was observed -- with sufficiently large samples total-separation partition indicates that there is no real relationship underlying the various clusterings.  
 
\begin{table}
\caption{Results of consensus clustering of randomly selected partitions with upper limit on $k$ equal 4}\label{tab:randomconsensusklimited}
\begin{tabular}{|p{2cm}|p{2cm}|p{2cm}|p{2cm}|p{2cm}|}
\hline
data set size & number of all possible clusterings 
& sample size & percentage of possible clusterings & percentage of total-separation partitions in consensus \\
\hline
\textit{1} & \textit{2} & \textit{3} & \textit{4} & \textit{5} \\
\hline
\hline
6 & 187 & 187 & 100 \% & 100 \%   \\ 
6 & 187 & 94 & 50 \% & 100 \%   \\ 
6 & 187 & 75 & 40 \% & 100 \%   \\ 
6 & 187 & 37 & 20 \% & 99.6 \%   \\ 
6 & 187 & 19 & 10 \% & 81.5 \%   \\ 
\hline
7 & 715 & 143 & 20 \% & 100 \%   \\ 
7 & 715 & 72 & 10 \% & 100 \%   \\ 
7 & 715 & 57 & 8 \% & 99.8 \%   \\ 
7 & 715 & 29 & 4 \% & 95.5 \%   \\ 
7 & 715 & 14 & 2 \% & 76.6 \%   \\ 
\hline
\end{tabular}
\end{table}

\inserted{
The question can be raised: Can it really be so bad that modern day clustering techniques would provide a clustering when no clusters are there in the data. 
Vast majority of clustering algorithms produce partitions whatever data they get. We can just point at $k$-means algorithm, but others, like DBSCAN, single-link etc. could be used. 
Imagine a large collection of data points in a high-dimensional space. 
Furthermore imagine the points are randomly uniformly distributed in space. 
Imagine that in order to perform $k$-means clustering efficiently, samples from this collection are drawn and $k$-means algorithm is performed on each of them. 
As $k$-means always clusters the entire space (computes centroids), we would have 
in fact a random sample of partitions from this universe. 
Now let the many partitions be processed by consensus clustering using Rand Index. 
A consensus partition will be undoubtedly found. 
The lesson from our theorem will be that it should converge towards total-separation partition. 

But we can face also another serious real problem.
Assume that we have a data set with an underlying structure (say with two dense areas separated by empty space), but located in a highly dimensional space.  
$k$-means algorithm has the known property of being $k$-rich that is upon proper transformation of distances between data points ANY clustering consisting of $k$ clusters may be obtained. And upon applying various clustering algorithms indeed  distances are transformed, e.g. by standardization, normalization, spectral transformations etc.  This means, however, that upon irresponsible choice of distance transformations we may sample from the universe of all possible partitions in spite of the fact that the data has originally a structure. And here again, as a result of consensus clustering, we can get  total-separation partition, which would be really bothering. So any transformation we apply to the data as an element of the clustering process must not violate the geometrical structure of the data space we expect to see.
}

\inserted{
So whether or not we restrict ourselves to an upper limit of clusters in a partition, 
same answers are to be given to our  questions driving the experiments.  
So either we have to live with the risk that a consensus clustering returns us 
a meaningless consensus or have to further develop these consensus methods so that they are able to refuse to return a partition if no real structure in the data exists. 
}

\section{Remarks on meta-clustering}\label{sec:metaclustering}

The formulas derived in the preceding sections shed also some light on possible outcome of 
meta-clustering. 
It is obvious that assuming a sufficiently "large" 
number of clusters, the simple extensions will be cluster centres and clusters will 
consist of extensions  of the same reduct.

Let us introduce the concept of p-th order reduct and p-th order extension.  
If  $\Gamma^*$ is a reduct of
$\Gamma$, then it is the 1st order reduct of $\Gamma$,
and $\Gamma$ is the 1st order extension of $\Gamma^*$.   
If it is a simple extension, then it is the 1st order simple extension. 
If  $\Gamma^+$ is a p-th order reduct of
$\Gamma^*$
and  $\Gamma^*$ is a reduct of
$\Gamma$, then  $\Gamma^+$ is the (p+1)st order reduct of $\Gamma$ and $\Gamma$ is the (p+1)st order extension of $\Gamma^+$.
If at the same time   $\Gamma^*$ is a p-th order simple extension  of
$\Gamma^+$
and  $\Gamma$ is a simple extension  of
$\Gamma^*$, then  $\Gamma$ is the (p+1)st order simple reduct of $\Gamma^+$. Otherwise $\Gamma$ is its complex extension. 
 
With the above definition let us   arrange a hierarchical clustering, where 
 the p-th level (from the bottom) consists of clusters with common p-th degree reduct.
In this case it is easy to see from  Theorem \ref{th:consensus}  that the cluster centres would be   p-th degree simple extensions of the respective reduct, around which the cluster is defined.
This hierarchy would be a local minimum for the combined distance of elements from their cluster centres at respective levels (each element of a cluster is closer to its own cluster centre than to any other cluster centre). 
One can in fact check that at the top level of such a hierarchy moving any object between classes is not possible. 

 Let us demonstrate it by considering two meta-cluster centres over the set of partitions of $n$ elements:
$\Gamma_0$ being total-separation partition and $\Gamma_1$ in which elements $1,2$ are in one cluster and the others \Bem{s inserted} in separate clusters. 
Consider now a partition $\Gamma_q$ in which elements 1 and 2 are in separate clusters, and no assumption with respect to other is done. 
\begin{align}
 unCD(\Gamma_0,\Gamma_q) 
  &= I_{12/0q}+\sum_{j=3}^{n }  I_{1j/0q}+\sum_{j=3}^{n }  I_{2j/0q}+
\sum_{i=3}^{n-1} \sum_{j=i+1}^{n }  I_{ij/0q}  
\nonumber \\
 unCD(\Gamma_1,\Gamma_q) 
  &= I_{12/1q}+\sum_{j=3}^{n }  I_{1j/1q}+\sum_{j=3}^{n }  I_{2j/1q}+
\sum_{i=3}^{n-1} \sum_{j=i+1}^{n }  I_{ij/1q}  
\end{align}
where $I_{jk/lm}$ means indicator of membership of element $j,k$  
in same cluster in one of partitions $\Gamma_l, \Gamma_m$  and in different in the other.
It is easily checked that
$\sum_{j=3}^{n }  I_{1j/0p}=\sum_{j=3}^{n }  I_{1j/1p}$, 
$\sum_{j=3}^{n }  I_{2j/1p}=\sum_{j=3}^{n }  I_{2j/1p}$
and 
$\sum_{i=3}^{n-1} \sum_{j=i+1}^{n }  I_{ij/0p}  =\sum_{i=3}^{n-1} \sum_{j=i+1}^{n }  I_{ij/1p}  $. 
The only difference is 
$ I_{12/1q}$ equal to one and $ I_{12/0q}$ equal to zero.
$\Gamma_q$ is closer to $\Gamma_0$ than to $\Gamma_1$, as expected.
If on the other hand both 1 and 2 would be in the same cluster in $\Gamma_q$, the situation would be inversed. 

Let us consider more detailed levels of meta-clustering, that is let $\Gamma_0$ and $\Gamma_1$ be pth simple extensions of 
$\Gamma^*_0$ and $\Gamma^*_1$ (over same set of elements) resp. and let $\Gamma_q$ be pth  extension  of 
$\Gamma^*_0$. 

\begin{align}
 unCD(\Gamma_0,\Gamma_q) 
\Bem{
&=\sum_{i=1}^{n-p} \sum_{j=i+1}^{n }  I_{ij/0q}  
 +\sum_{i=n-p+1}^{n-1} \sum_{j=i+1}^{n }  I_{ij/0q}  
}
&=\sum_{i=1}^{n-p} \sum_{j=i+1}^{n-p }  I_{ij/0q}  
 +\sum_{i=1}^{n-p} \sum_{j=n-p+1}^{n }  I_{ij/0q}  
 +\sum_{i=n-p+1}^{n-1} \sum_{j=i+1}^{n }  I_{ij/0q}  
\nonumber \\
 unCD(\Gamma_1,\Gamma_q) 
&=\sum_{i=1}^{n-p} \sum_{j=i+1}^{n-p }  I_{ij/1q}  
 +\sum_{i=1}^{n-p} \sum_{j=n-p+1}^{n }  I_{ij/1q}  
 +\sum_{i=n-p+1}^{n-1} \sum_{j=i+1}^{n }  I_{ij/1q}  
\end{align}
Again obviously 
 $\sum_{i=1}^{n-p} \sum_{j=n-p+1}^{n }  I_{ij/0q}  
 =\sum_{i=1}^{n-p} \sum_{j=n-p+1}^{n }  I_{ij/1q}  $,
 $\sum_{i=n-p+1}^{n-1} \sum_{j=i+1}^{n }  I_{ij/0q}  
 =\sum_{i=n-p+1}^{n-1} \sum_{j=i+1}^{n }  I_{ij/1q}  $,
so that again  
$\sum_{i=1}^{n-p} \sum_{j=i+1}^{n-p }  I_{ij/0q}=0$  
(as both $\Gamma_0$ and $\Gamma_q$ have the same reduct)   
and
$\sum_{i=1}^{n-p} \sum_{j=i+1}^{n-p }  I_{ij/1q}  >0$
make the difference. 
One concludes that if the meta-clustering is a hierarchical one and the clusters are build around same reducts then the clusters at each hierarchy level are stable.

Hence it is obvious that also meta-clustering is biased - there exists a structure in spite of filling in the whole sample space.

The above meta-clustering "algorithm" was pretty much manual. 
One can ask whether or not other algorithms will exhibit the same tendency.
In particular if it turns out that the total-separation partition is chosen as centre of any of the meta-clusters.

  Let us try out  $k$-medoids clustering, implemented in $R$ as $pam$ algorithm and the popular  $k$-\emph{means} algorithm as meta-clustering methods. 
For this purpose let us span 
a $(n-1)\cdot n/2$ dimensional space for partitions of $n$ elements.
For a partition $\Gamma$, for $1\le i<j\le n$ the $(j-1)\cdot(j-2)/2+i$-th coordinate in this space 
would be equal to 1 if both elements $i,j$ belong to the same cluster and  equal to 0 otherwise.
It is easily seen that $unCD$ is the taxicab-distance between partitions in this space  or the square of Euclidean distance, making it reasonable to apply e.g. $k$-\emph{means} algorithm.
Let us have a look at the  meta-clusters   generated  using $k$-\emph{means} and $pam$ algorithm (as implemented in $R$ system)  under this representation.

For a set of 4 elements, 
$k$-\emph{means} algorithm with $k=2$  splits the set of all partitions into the meta-clusters:
 Meta-cluster  1 with minimal distance to meta-cluster centre  1.5 containing 10 
partitions:   
\\ $\{   \{ 1, 2, 4 \}  \{ 3 \} \},   $
\\ $\{ 1, 2 \}  \{ 3, 4 \}  \},$
\\ $\{   \{ 1, 2 \}  \{ 3 \}  \{ 4 \} \},$
\\ $\{   \{ 1, 4 \}  \{ 2, 3 \} \},$
\\ $\{    \{ 1 \}  \{ 2, 3, 4 \} \},$
\\ $\{   \{ 1 \}  \{ 2, 3 \}  \{ 4 \} \},$
\\ $\{	   \{ 1, 4 \}  \{ 2 \}  \{ 3 \} \},$
\\ $\{	   \{ 1 \}  \{ 2, 4 \}  \{ 3 \} \},$
\\ $\{	  \{ 1 \}  \{ 2 \}  \{ 3, 4 \} \},$
\\ $\{	  \{ 1 \}  \{ 2 \}  \{ 3 \}  \{ 4 \} \}. $
\\The last one   is the closest point to centroid of this meta-cluster. 
 Meta-cluster  2 with minimal distance to meta-cluster centre  2 containing 5 
partitions:   
\\ $\{   \{ 1, 2, 3, 4 \} \},$
\\ $\{	 \{ 1, 2, 3 \}  \{ 4 \} \},$
\\ $\{	 \{ 1, 3, 4 \}  \{ 2 \} \},$
\\ $\{	 \{ 1, 3 \}  \{ 2, 4 \} \},$
\\ $\{	 \{ 1, 3 \}  \{ 2 \}  \{ 4 \} \}$
\\The last one   is the closest point to centroid of this meta-cluster. 

On the other hand, the $pam$ algorithm with split parameter set to 2
creates a meta-cluster 
consisting of 
partitions
$   \{ \{ 1, 2, 3, 4 \}\}$
and  $\{   \{ 1 \}  \{ 2, 3, 4 \}\}$, while the other meta-cluster contains 
all the other partitions, with 
 $\{  \{ 1 \}  \{ 2 \}  \{ 3 \}  \{ 4 \} \}$ (total-separation partition) being the medoid.

For 5 elements $k$-\emph{means} provides two meta-clusters:
meta-cluster  1 minimum distance to meta-cluster centroid equal  2.432 
and of  card. 37
with element closest to the centroid $\{ \{ 1 \}  \{ 2 \}  \{ 3 \}  \{ 4 \}  \{ 5 \} \} $
and a meta-cluster   2 (min. dist. 3 card. 15)
with element closest to the centroid
 $\{ \{ 1, 2 \}  \{ 3 \}  \{ 4 \}  \{ 5 \} \}$. 

The two meta-clusters returned by $pam$
have cardinality   6 with medoid 
  $\{\{ 1, 2, 3, 4, 5 \}\}$ and   card. 46 with medoid 
 $\{ \{ 1, 2 \}  \{ 3 \}  \{ 4 \}  \{ 5 \} \}$. 
Here, total-separation partition was not selected, but still is among the candidates. 
 
For 6 elements $k$-\emph{means} creates one meta-cluster of card. 156 with the total-separation partition being the closest one to the centroid, while the other meta-cluster 
of card. 47 
has 25 elements with minimal distance of  7.40 from the centroid.

$pam$ provides two meta-clusters of card. 52 and 151 with medoids 
$\{ \{ 1 \}  \{ 2 \}  \{ 3 \}  \{ 4 \}  \{ 5, 6 \} \}$ and 
total-separation one resp.

For 7 elements $k$-\emph{means} returns a meta-cluster of card. 582 with the total-separation partition being the closest one (dist. 3.463) to the centroid, 
and another meta-cluster of card. 295
105 elements closest to the centroid (dist. 9.264406)

$pam$ provides two meta-clusters of card. 203 and 674 with medoids 
$\{ \{ 1 \}  \{ 2 \}  \{ 3 \}  \{ 4 \}  \{ 5 \} \{ 6, 7 \} \}$ and 
total-separation one resp.

For 8 elements in $k$-\emph{means} we have 
 one meta-cluster of card. 2570 with the total-separation partition being the closest one to the centroid (dist. 4.21), while the other meta-cluster 
of card. 1570 
has 700 elements with minimal distance of  11.37 from the centroid.  
 
$pam$ provides two meta-clusters of card. 877 and 3263 with medoids 
$\{ \{ 1 \}  \{ 2 \}  \{ 3 \}  \{ 4 \}  \{ 5 \}  \{ 6 \} \{ 7, 8 \} \}$ and 
total-separation one resp. 
 
For $k$-\emph{means} it should be underlined that the distance between cluster centres does not exceed 1 in any of the above cases.

It is worth noting  that the meta-cluster around the total-separation partition 
is the larger one. 
In fact, with $k$-\emph{means} 
for 8 elements if $k$ grows even to 100,
this meta-cluster (with total-separation partition) is the largest one. 
The percentage of variance explained (betweenness/ totals) 
is however low. 
With k=2  3\% is explained,
with k=20 26\% is explained,
with k=50 41\% is explained, 
with k=100 54\%, with k=150 60\%, with k=200 64\%. 

We also see in these experiments that split into two classes as performed by $pam$ is in agreement with our claim that p-th order simple extensions will become meta-cluster centres. 
This is even true for $k$-\emph{means} with the total separation partition. 

A comment on the difference between $pam$ and $k$-\emph{means} results seems to be required. 
First of all we made here a trick of pressing the distances between partitions into a vector space.
This should not affect $pam$ algorithm as the distances in this space are the same as the original space, but it may be a little bit confusing for $k$-\emph{means} which actually uses the square roots of the original distances. 
Furthermore, the dense partition space was replaced by a sparse vector space. This leads $k$-\emph{means} to explore places in this space (as centroids) that also not have a clear interpretation. Possibly one could investigate a kind of fuzzification here in order to have an insight into what the centroid may mean. This may be a future research path.

Finally, let us have a look at the performance of $pam$ when we do not (meta)-cluster the universe of all possible partitions, but only random samples from it. 
We check again if the total-separation partition is among the candidates. 
We have, however, one difference here compared to the settings of experiments in Section \ref{sec:consensus-implications}. As $pam$ seeks for medoids, total-separation partition must be among the partitions to be meta-clustered, so we enforce selecting it when randomly sampling. 
Table \ref{tab:randommeta} shows the behaviour of $pam$ when the samples are uniformly drawn from the universe of partitions.  Total-separation partition is the dominating option for medoid of at least one of two meta-clusters, that $pam$ is requested to generate. 
Table \ref{tab:randommetastruct} shows the behaviour of $pam$ when we draw the samples from the sub-universe with a structure as done in Table \ref{tab:structconsensus}.   
Here, for sufficiently large random samples the total-separation partition, though present in each sample, is rarely chosen.

\begin{table}
\caption{Results of meta-clustering via $pam$ of randomly selected partitions}\label{tab:randommeta}
\begin{tabular}{|p{2cm}|p{2cm}|p{2cm}|p{2cm}|p{2cm}|}
\hline
data set size & number of all possible clusterings 
& sample size & percentage of possible clusterings & percentage of total-separation partitions as meta-cluster centre \\
\hline
\textit{1} & \textit{2} & \textit{3} & \textit{4} & \textit{5} \\
\hline
\hline
4 & 15 & 15 & 100 \% & 100 \% \\ 
4 & 15 & 8 & 50 \% & 64.9 \% \\ 
\hline
5 & 52 & 52 & 100 \% & 100 \% \\ 
5 & 52 & 27 & 50 \% & 99.6 \% \\ 
5 & 52 & 21 & 40 \% & 97.8 \% \\ 
5 & 52 & 11 & 20 \% & 83.7 \% \\ 
5 & 52 & 6 & 10 \% & 68.9 \% \\ 
\hline
6 & 203 & 203 & 100 \% & 100 \% \\ 
6 & 203 & 102 & 50 \% & 100 \% \\ 
6 & 203 & 82 & 40 \% & 100 \% \\ 
6 & 203 & 41 & 20 \% & 99.9 \% \\ 
6 & 203 & 21 & 10 \% & 98 \% \\ 
\hline
7 & 877 & 89 & 10 \% & 100 \% \\ 
7 & 877 & 45 & 5 \% & 100 \% \\ 
7 & 877 & 36 & 4 \% & 100 \% \\ 
7 & 877 & 19 & 2 \% & 99.6 \% \\ 
7 & 877 & 10 & 1 \% & 96.9 \% \\ 
\hline
8 & 4140 & 84 & 2 \% & 100 \% \\ 
8 & 4140 & 42 & 1 \% & 100 \% \\ 
8 & 4140 & 34 & 0.8 \% & 100 \% \\ 
8 & 4140 & 18 & 0.4 \% & 99 \% \\ 
8 & 4140 & 9 & 0.2 \% & 99 \% \\

\hline
\end{tabular}
\end{table}

\begin{table}
\caption{Results of meta-clustering via $pam$ of randomly selected partitions out of a set with structure}\label{tab:randommetastruct}
\begin{tabular}{|p{2cm}|p{2cm}|p{2cm}|p{2cm}|p{2cm}|}
\hline
data set size & number of all possible clusterings 
& sample size & percentage of possible clusterings & percentage of total-separation partitions as meta-cluster centre \\
\hline
\textit{1} & \textit{2} & \textit{3} & \textit{4} & \textit{5} \\
\hline
\hline
4 & 6 & 6 & 100 \% & 0 \% \\ 
4 & 6 & 3 & 50 \% & 10.8 \% \\ 
\hline
5 & 16 & 16 & 100 \% & 0 \% \\ 
5 & 16 & 9 & 50 \% & 0.2 \% \\ 
5 & 16 & 7 & 40 \% & 1.4 \% \\ 
5 & 16 & 4 & 20 \% & 10.1 \% \\ 
5 & 16 & 3 & 10 \% & 25 \% \\ 
\hline
6 & 53 & 53 & 100 \% & 0 \% \\ 
6 & 53 & 27 & 50 \% & 0 \% \\ 
6 & 53 & 22 & 40 \% & 0 \% \\ 
6 & 53 & 11 & 20 \% & 1.5 \% \\ 
6 & 53 & 6 & 10 \% & 9.5 \% \\ 
\hline
7 & 204 & 42 & 20 \% & 0.1 \% \\ 
7 & 204 & 21 & 10 \% & 3.6 \% \\ 
7 & 204 & 17 & 8 \% & 5.1 \% \\ 
7 & 204 & 9 & 4 \% & 19.6 \% \\ 
7 & 204 & 5 & 2 \% & 35.2 \% \\ 
\hline
8 & 878 & 71 & 8 \% & 6 \% \\ 
8 & 878 & 36 & 4 \% & 16 \% \\ 
8 & 878 & 29 & 3.2 \% & 22 \% \\ 
8 & 878 & 15 & 1.6 \% & 32 \% \\ 
8 & 878 & 8 & 0.8 \% & 43 \% \\

\hline
\end{tabular}
\end{table}

\section{Conclusions and future work} \label{sec:conclusions}

As the number of available clustering algorithms applicable to the same data is growing, and the potential outputs may differ substantially, methodologies to reconcile them like meta-clustering or consensus clustering are under development.

In this paper we demonstrated 
that both consensus clustering and meta-clustering 
using Cluster Difference (derived from Rand Index) as a measure of distance 
between partitions,
when applied to the universe of all possible partitions, 
point to the partition containing each element in a separate set
as the best compromise. 

\Bem{
It is quite easy to invent clustering algorithms delivering for the same set of data any clustering we want.
But in the space of all partitions we get lost both by meta-clustering and consensus clustering.
Because meta-clustering will provide us with a structure of partitions that has nothing to do with the data
and consensus clustering will deliver the most trivial consensus having nothing to do with the data. 
}

This suggests that the user performing the task of clustering must at least have an approximate  vision of the geometry of the data space.
Only in this case the mentioned techniques may be helpful in the choice of appropriate compromise clustering. 

It seems also worth investigating, how other cluster quality functions used as distances between partitions would behave under consensus clustering of the space of all possible partitions.

It seems also worth investigating how such measures would behave 
not in the full universe of all partitions but rather for uniform random samples of it.  
Such a sampling would then constitute a background for investigations into the behaviour  of other partition comparison indexes, of consensus and meta-clustering methods as well as for checking if a resultant consensus-partition or meta-cluster really gives a new insight or is just a random artefact.

\Bem{
Therefore it is necessary to shift our attention from developing clustering algorithms as such 
to the investigation whether or not the clustering algorithm discovers something in the data or not, and if it discovers then how the discovered clusters are defined.

In particular one needs to establish new methods of comparing partitions in such a way as not to impose any bias onto the space of all partitions. Or, if we accept consciously the bias, it needs to have a justification drawn from outside, e.g. business benefit etc. 
This is an important research direction for the future. 

We drew our conclusions in this paper based on an investigation of the entire space of partitions. Therefore we were limited in fact to really small sets of elements. But we can guess that a "uniform" sampling of this universe would lead to sets of partitions exhibiting similar behaviour. One needs to verify this hypothesis and to derive an appropriate sampling method, which would be computationally feasible for sets of objects of practical size.
Such a sampling would then constitute a background for investigations into the biases of other partition comparison indexes, of consensus and meta-clustering methods as well as for checking if a resultant consensus-partition or meta-cluster really gives a new insight or is just a random artefact.  
}
 
\section*{Acknowledgements}
The author wishes to thank to the Institute of Computer Science of Polish Academy of Sciences for promoting and financing this research.

\bibliographystyle{fundam}

\bibliography{podobienvstwoMAK_bib}

\end{document}